\documentclass[conference]{IEEEtran}
\IEEEoverridecommandlockouts
\usepackage{cite}
\usepackage{amsmath,amssymb,amsfonts}
\usepackage{algorithmic}
\usepackage{graphicx}
\usepackage{textcomp}
\usepackage{xcolor}
\def\BibTeX{{\rm B\kern-.05em{\sc i\kern-.025em b}\kern-.08em
		T\kern-.1667em\lower.7ex\hbox{E}\kern-.125emX}}
\begin{document}
	
	\title{Leveraging Online Data to Enhance Medical Knowledge in a Small Persian Language Model}
	
	\author{\IEEEauthorblockN{1\textsuperscript{st} Mehrdad Ghassabi}
		\IEEEauthorblockA{\textit{Faculty of Computer Engineering} \\
			\textit{University of Isfahan}\\
			Isfahan, Iran \\
			m.ghassabi@eng.ui.ac.ir}
		\and
		\IEEEauthorblockN{2\textsuperscript{nd} Pedram Rostami}
		\IEEEauthorblockA{\textit{School of Electrical and Computer Engineering} \\
			\textit{University of Tehran}\\
			Tehran, Iran \\
			pedram.rostami@ut.ac.ir}
		\and
		\IEEEauthorblockN{3\textsuperscript{rd} Hamidreza Baradaran Kashani}
		\IEEEauthorblockA{\textit{Faculty of Computer Engineering} \\
			\textit{University of Isfahan}\\
			Isfahan, Iran \\
			hrb.kashani@eng.ui.ac.ir }
		\and
		\IEEEauthorblockN{4\textsuperscript{th} Amirhossein Poursina}
		\IEEEauthorblockA{\textit{School of Medicine} \\
			\textit{Isfahan University of Medical Sciences}\\
			Isfahan, Iran \\
			Amirhosseinpoorsina9@gmail.com}
		\and
		\IEEEauthorblockN{5\textsuperscript{th} Zahra Kazemi}
		\IEEEauthorblockA{\textit{Faculty of Computer Engineering} \\
			\textit{University of Isfahan}\\
			Isfahan, Iran \\
			zhrakazemi@mehr.ui.ac.ir}
		\and
		\IEEEauthorblockN{6\textsuperscript{th} Milad Tavakoli}
		\IEEEauthorblockA{\textit{Faculty of Computer Engineering} \\
			\textit{University of Isfahan}\\
			Isfahan, Iran \\
			m.tavakoli@mehr.ui.ac.ir}
	}
	
	\maketitle
	
	\begin{abstract}
The rapid advancement of language models has demonstrated the potential of artificial intelligence in the healthcare industry. However, small language models struggle with specialized domains in low-resource languages like Persian. While numerous medical-domain websites exist in Persian, no curated dataset or corpus has been available—making ours the first of its kind. This study introduces a newly curated dataset comprising 20k doctor-patient Q\&A pairs and 60\% of a 90-million-token crawled corpus from medical magazines. Using a parameter-efficient fine-tuning approach, we enhanced the medical knowledge of the baseline model, aya-expanse-8b. Benchmark evaluations demonstrate that the fine-tuned model achieves improved accuracy in medical question answering and successfully passed the Iranian Basic Medical Science Entrance Exam (IBSEE) in September 2023, which the baseline model did not. Additionally, the fine-tuned model improved Persian-translated MMLU accuracy by an average of 2.67\%. This work highlights the potential of leveraging open-access online data to enrich small language models in medical fields, providing a novel solution for Persian medical AI applications suitable for resource-constrained environments. Future research could explore multimodal input to further enhance performance.
	\end{abstract}
	
	
	\section{Introduction}
	The advent of the transformer architecture, as introduced in the groundbreaking paper “Attention is All You Need”
	\cite{b1}
	has catalyzed a rapid evolution in the field of natural language processing (NLP). This innovation has led to the development of increasingly sophisticated language models that leverage attention mechanisms to understand and generate human language with remarkable accuracy. As a result, the integration of artificial intelligence (AI) into various domains has surged, particularly in the medical field, where AI-driven solutions are being employed to enhance diagnostic accuracy, patient care, and administrative efficiency.
	
	Despite the vast amount of research and development dedicated to English medical language models, such as Med-Palm
	\cite{b2} \cite{b3}
	and others, there remains a significant disparity in resources available for non-English languages, particularly Persian. To the best of our knowledge, the only existing Persian medical language model, Sina-BERT
	\cite{b4}
	, is a closed-source solution, limiting its accessibility and adaptability for further research and application. This gap underscores the urgent need for open-source resources that cater specifically to the Persian-speaking medical community.
	
	However, this gap stems from underutilization rather than a lack of raw material. Persian-language medical forums (e.g., drhast,doctor-yab) and authoritative online magazines (e.g., hidocor, niniban) host vast amounts of expert-curated content and real-world patient-doctor interactions. These sources—if systematically crawled, cleaned, and structured—could serve as valuable resources for training a domain-specific Persian language model.
	
	Moreover, the development of small language models is particularly crucial in the medical domain due to privacy concerns. These models can be optimized to run on local devices, ensuring that sensitive patient data remains secure and confidential, which is a paramount consideration in healthcare settings \cite{b5}.
 However, the unavailability of appropriate medical corpora and datasets in Persian has hindered progress in this area, impeding the creation of robust language models that can effectively address the linguistic and cultural nuances of the Persian-speaking population.
	
	In response to these challenges, we present a novel approach with our model, Gaokerena-V
	\footnote{
		Our language model is named after Gaokerena, an ancient Persian mythological tree believed to possess healing properties and grant immortality to those who consume its fruit.
	}
	, which fine-tunes a baseline model, aya-expanse-8b \cite{b6}, on a crawled dataset comprising a Persian medical corpus and a newly introduced free-form Farsi medical question-answering dataset. The crawled corpus contained 90 million tokens, of which 60\% was utilized for training the model. Additionally, we introduced a new dataset named MF3QA, which contains 20,000 patient-doctor interaction pairs, further enhancing the model’s ability to handle diverse medical queries. aya-expanse-8b was specifically chosen as the baseline model due to its strong understanding of Persian language grammar, making it an ideal foundation for a Persian language model. Importantly, our model, corpus, and datasets are all open-source, promoting transparency and collaboration within the research community. This development aims to enhance access to Persian medical information and support secure, efficient interactions within the healthcare environment. By bridging the existing gaps in resources and leveraging advancements in NLP, our work contributes to the growing landscape of AI in medicine, particularly for Persian-speaking users.

	Our contributions in this work are as follows:
	\begin{itemize}
		\item Introducing the first open-source
		\footnote{
			we have published our model and datasets at https://huggingface.co/gaokerena
		}
		persian medical language model that achieved promissing result in
		comparison to other home device runnable alternatives
		\item Introducing a persian medical corpus obtained by crawling different websites.
		\item  Introducing the first persian free form medical question answering dataset obtained by crawling different websites.
		\item translating medical portion of MMLU benchmark which can be used to evaluate any persian medical language model
	\end{itemize}
	
	\section{Related Work}
	
	\subsection{Related Works in English}
	Several notable projects have contributed to the development of medical language models, employing various strategies to enhance their performance and applicability in healthcare.
	
	ChatDoctor
	\cite{b7}, which is the most similar work to ours, represents a notable initiative focused on developing a medical language model. The team behind ChatDoctor sourced its training data from HealthcareMagic and its test data from iCliniq, compiling a total of more than 200,000 free-form question-answering pairs from these online platforms. They then curated the dataset by filtering answers based on their length, resulting in a final collection of 100,000 high-quality pairs. Using this dataset, they fine-tuned a LLaMA model \cite{b8} to create a system capable of delivering accurate and contextually relevant medical information. Furthermore, ChatDoctor leveraged a retrieval-augmented generation (RAG) approach, which allowed the model to access and integrate external knowledge more effectively, thereby enhancing its overall performance.
	
	Meerkat
	\cite{b9}
	is another significant contribution in this field. This project involved extracting chains of thought from medical textbooks and fine-tuning a language model using this data, alongside other supplementary datasets. By focusing on the reasoning processes involved in medical decision-making, Meerkat aimed to create a model that not only provides information but also mimics the cognitive processes of healthcare professionals, thereby supporting more nuanced and informed interactions.
	
	MedMobile
	\cite{b10}
	represents yet another advancement in the realm of small medical language models. This work fine-tuned the Phi-3-mini model
	\cite{b11}
	using a combination of synthetic and human-generated datasets, enabling it to achieve optimal performance tailored for mobile applications in the medical domain. By focusing on the specific requirements of mobile users, MedMobile sought to deliver a model that is both efficient and effective, ensuring accessibility to high-quality medical information on the go.
	\subsection{Related Works in Persian}
As previously mentioned, there has been limited research focused on Persian medical language models, highlighting a significant gap in resources for the Persian-speaking medical community. Furthermore, existing works on Persian medical question-answering systems are entirely closed-source regarding their datasets, models, and codebases. This lack of public resources leaves the field largely underexplored, presenting researchers with an almost blank slate to build upon. On the other hand, all of these efforts have primarily concentrated on extractive solutions, which aim to retrieve relevant information from predefined sources, rather than employing generative approaches capable of producing context-aware responses. In contrast to all previous studies in the Persian language domain, our approach is generative, not extractive, leveraging the internal knowledge of a language model to produce answers by synthesizing context and understanding. Generative question-answering systems rely on the internal knowledge of a language model, enabling them to produce answers by synthesizing context and understanding, whereas extractive question-answering systems retrieve answers from expert-generated data. For a detailed comparison of these two approaches, please refer to Table \ref{tab:ext_gen_comparison}.

	\begin{table}[ht]
		\centering
		\caption{Comparison of extractive question answering systems and  generative question answering systems.} 
		\begin{tabular}{|l|c|c|}  
			\hline
			& extractive QA systems & generative QA systems \\ \hline
			relies on & information retrieval & generative AI \\ \hline
			using LMs & encoder based LMs & decoder based \&   \\ 
			 &  & encoder-decoder based LMs  \\ \hline
                        architecture & expert generated data +  & a generative \\ 
                         & an information &  language model \\ 
                         & retrieval system &  \\ \hline
			advantages& reliability  &  flexibility  \\ \hline
			disadvantages& data dependency  & hallucination	\\ \hline
		\end{tabular}
		\label{tab:ext_gen_comparison}
	\end{table}
	
	Perhaps the most notable effort in the Persian medical language models field is Sina-BERT \cite{b4}, which involved training a BERT model \cite{b12}
	 using a crawled corpus alongside Persian annotated datasets specifically developed for various tasks, including medical question answering, medical sentiment analysis, and medical question retrieval. Sina-BERT is the most similar work to ours among Persian language focused efforts; however, it differs in that it uses a BERT model—an encoder-based language model—as its baseline. This choice limits its capability for generative AI tasks, as BERT is primarily designed for understanding and extracting information rather than generating answers.
	
	Another notable work is the Persian Medical Question Answering System developed by H. Veisi et al.
	\cite{b13}.
	Their system is structured around three main modules: question processing, document retrieval, and answer extraction. The question processing module is responsible for analyzing and refining user queries, the document retrieval module locates relevant medical documents from predefined data, and the answer extraction module identifies and extracts the most suitable answers from the retrieved content.
	
	Similar to these two works L.Darabi
	\cite{b14}
	used models like Pars-BERT
	\cite{b15}
	to retrieve relevant answers. Her approach involves finding similar questions to handle repeated queries and employs strict and lenient evaluation strategies for accurate or approximate answers. Additionally, classification methods and Named Entity Recognition (NER) are used to improve answer relevance by categorizing questions and identifying medical entities like drug and disease names.
	
	\section{Baseline Model}
	We have chosen aya-expanse-8b as our baseline model primarily due to the lack of open-source Persian medical language models, which necessitates the use of a general-purpose language model. While there are several multilingual options available, including aya-expanse, gemma-2 \cite{b16}, qwen2.5 \cite{b17}, and persianmind \cite{b18}, we have determined that aya-expanse is the most suitable choice for our needs. One key reason is that the training data for the other models predominantly consists of non-Persian languages, leading to biases that may result in the generation of non-Persian characters, even when we explicitly instruct the model to use only Persian. In contrast, aya-expanse demonstrates a robust understanding of Persian grammar and produces grammatically rich Persian text, making it a more reliable option for our research. Furthermore, if we merge our updated parameters into aya-vision \cite{b19}, another model from the aya family, instead of aya-expanse, we gain the capability to incorporate medical images such as MRIs and CT scans as inputs, thereby enhancing our model’s ability to process complex medical questions.
	\section{Data}
	
	\subsection{Corpus}
	As previously mentioned, there is a notable absence of publicly available Persian medical corpora specifically collected for training machine learning models. This lack of a dedicated Persian medical corpus poses a significant challenge for researchers and developers aiming to create effective models for medical applications in the Persian language. Without high-quality, domain-specific textual data necessary for training, these efforts may be hindered, ultimately impacting the development of advanced medical technologies and solutions tailored for Persian-speaking populations. To provide further insight into this issue, we have compiled a comprehensive corpus containing approximately 90 million tokens and about 100,000 articles. 
	
\begin{figure}[htbp]
		\centerline{\includegraphics[width=0.4\textwidth]{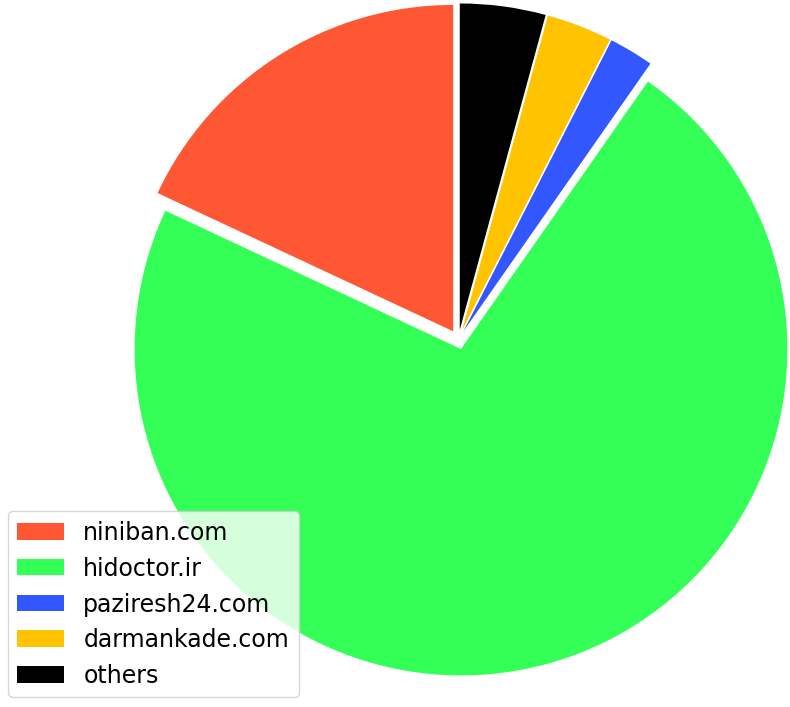}}
		\caption{Percentage of content from each magazine crawled for the corpus}
		\label{fig1}
	\end{figure}

	I. Garcia Ferrero et al.
	\cite{b20}
	collected medical corpora dedicated to four languages (English, French, Spanish, and Italian), which can be compared to ours, as shown in Table 
	\ref{tab:corpus_comparison}.
	The accompanying Figure \ref{fig1} illustrates the share of each magazine within our corpus, effectively highlighting the diversity of sources and underscoring the need to address gaps in available resources to foster innovation and improve health-related applications.
	
	\begin{table}[ht]
		\centering
		\caption{Comparison of Our Corpus with Corpora Collected by I. Garcia Ferrero et al.}
		\begin{tabular}{|l|c|c|}  
			\hline
			language& no. tokens & collected by \\ \hline
			English & 1.1B & I. Garcia Ferrero et al. \\ \hline
			Spanish & 950M & I. Garcia Ferrero et al.  \\ \hline
			French & 675M & I. Garcia Ferrero et al.  \\ \hline
			Italian& 143M &  I. Garcia Ferrero et al.  \\ \hline
			Persian& 90M & us	\\ \hline
		\end{tabular}
		\label{tab:corpus_comparison}
	\end{table}
	
	\subsection{Dataset}
	The collection of a real-world doctor-patient question-answering dataset is crucial for enhancing the capabilities of language models in the healthcare domain. Such a dataset allows models to learn valuable information derived from authentic interactions between healthcare providers and patients. By analyzing these real-world exchanges, language models can grasp the nuances of medical terminology, patient concerns, and the context surrounding healthcare inquiries. Furthermore, this dataset equips models with the ability to learn not just the factual content of responses but also the appropriate structure and tone for answering questions. This dual learning process is essential, as it enables the model to generate accurate, empathetic, and contextually relevant responses, ultimately improving patient communication and support in medical environments. In this context, Yang Liu 
	\cite{b21}
	highlights several real-world doctor-patient question-answering datasets in his survey, a comparison of these datasets with ours can be found in Table 
	\ref{tab:mf3qa_comparisin}. In an era where technology increasingly aids healthcare, a robust doctor-patient dataset stands as a foundational element in training models that can effectively contribute to better healthcare delivery.
	
	\begin{table}[ht]
		\centering
		\caption{Comparison of Our dataset with others}
		\begin{tabular}{|l|c|c|c|}  
			\hline
			dataset name            &language & no. records & collected by \\ \hline
			ChatDoctor 	   &English	 & 100K        & Yunxiang Li et al.
			\cite{b7}
			\\ \hline
			CMtMedQA  	   &Chinese	 & 68K        &  Songhua Yang et al.
			\cite{b22}
			\\ \hline
			DISC-Med	   &Chinese	 & 465K       & Zhijie Bao et al.   
			\cite{b23}
			\\ 
			-SFT       & &  &  \\ \hline
			HuatuoGPT-	   &Chinese	 & 226K        & Hongbo Zhang et al.   
			\cite{b24}
			\\ 
			sft-data-v1	   &	     &             & 	\\ \hline
			Huatuo-26M	   &Chinese	 & 26M         & Jianquan Li et al.
			\cite{b25}
			\\ \hline
			MedDialog       &Chinese \&&3.66M       & Guangtao Zeng et al.  \\ 
			&English  &             &  \cite{b26} \\ \hline
			Medical-       & &  &  \\
			Meadow  &English  & 160k        & Tianyu Han et al. 
			\cite{b27}
			\\ \hline
			MF3QA           &Persian  & 20k            & us \\ \hline
		\end{tabular}
		\label{tab:mf3qa_comparisin}
	\end{table}
	
In our research, we crawled more than 180,000 question-answer pairs from Persian medical forums using the BeautifulSoup library
\footnote{you can see the crawling codes at https://github.com/Mehrdadghassabi/Gaokerena-V/tree/main/dataset/MF3QA/crawling\_and\_filtering}
, employing both manual and automatic filtering methods to refine the dataset through a laborious cleaning process. This approach is similar to the work done by Yunxiang Li et al. \cite{b7}. In their article on the Chat Doctor medical language model, where they extracted data from English medical forums, Yunxiang Li discarded about half of the question-answer pairs based solely on the length of the answers, as shorter responses are generally inadequate for training a model and can lead the model to learn to provide brief answers. However, we faced a greater challenge; Persian doctors tend to provide much shorter answers compared to their English counterparts, resulting in the necessity to discard over 80\% of our question-answer records to ensure quality and relevance for our training purposes.

While Yunxiang Li only cleaned his question-answer pairs based on answer length, we used a more sophisticated process to do so. Firstly, we omitted answers with short lengths (less than 50 tokens) randomly in the training and development sets to shift the token-count histogram to the right. This random omission was not applied to the test set; instead, we meticulously reviewed each question-answer pair in the test set to ensure validity and accuracy. Additionally, we carefully read all question-answer pairs to filter out spam, duplicates, or low-quality records, ensuring that all remaining data contained high-quality information. We also removed any personal data to ensure compliance with privacy standards.

\begin{figure}[htbp]
		\centerline{\includegraphics[width=0.45\textwidth]{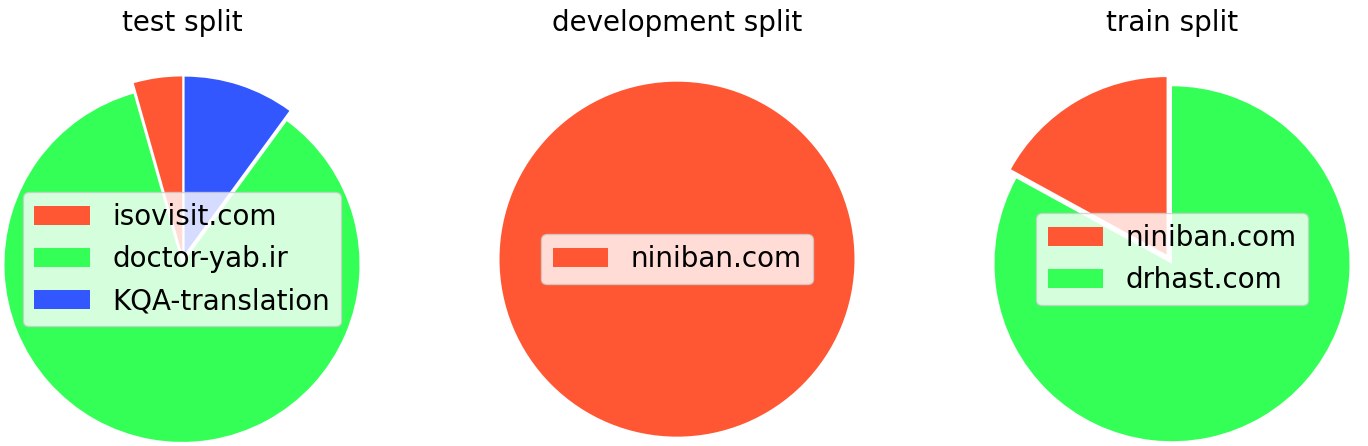}}
		\caption{Percentage of content from each forums crawled for MF3QA dataset}
		\label{fig2}
	\end{figure}

	As you can see in the Figure
	\ref{fig2}
	to create our dataset, we utilized patient-doctor interactions from the drhast and niniban platforms for the training split.It is important to note that drhast does not provide all of its doctor-patient interaction records on its site; it only offers access to the last 2,000 records. Additionally, each record is linked to 100 related records, complicating the crawling process. To address this challenge, we treated their data as a graph and performed a breadth-first search (BFS), which took about two weeks to extract 120,000 records out of a total of 200,000. The BFS we performed involved using a “for” loop where we selected one of the available 2,000 records as the root node and iteratively explored its linked records. by repeatedly performing BFS with new roots, we ensured broader coverage of the dataset. To prevent redundancy, we filtered out duplicated crawled records after each iteration. It is worth noting that the last 2,000 records on drhast are dynamic and change over time as new questions are asked, allowing us to use newly added records as roots for subsequent BFS runs. This iterative process, while time-consuming, ensured we could extract a substantial portion of the dataset within the two-week period. For the test set, we used the doctor-yab and isovisit sites, ensuring diversity by translating the K-QA question-answering dataset.
	\cite{b28}
	and appending it to our test split. This comprehensive approach not only enriched our dataset but also underscored the importance of real-world doctor-patient interactions in training effective language models.
	
	\section{Training}
	\subsection{Fine tuning}
We fine-tuned the 8-billion-parameter checkpoint of the aya-expanse model on 60\% of our corpus, focusing on minimizing resource usage.This decision was driven by the need to decrease the cost of training, as this project was entirely self-funded and constrained by a limited budget. To ensure an efficient fine-tuning process, we employed gradient checkpointing and a small batch size of 2,reducing memory requirements during training. Additionally,we used gradient accumulation steps of 16, effectively increasing the overall batch size to 32 and enabling stable training dynamics.

To further reduce the memory usage of our fine-tuning process, we leveraged Low Rank Adaptation (LoRA) \cite{b29} to significantly reduce the number of trainable parameters. Specifically, we implemented a rank of 8, an alpha value of 16, and a dropout rate of 5\%, applying the LoRA weights to all trainable parameters of each transformer layer.To further optimize the process, we employed efficient tokenization and memory-aware training techniques. The tokenization process divided the input text into manageable token sequences, ensuring consistent input and label structures by truncating, padding, and handling overflowed tokens to maintain contextual integrity within a fixed context length.

\begin{figure}[h]
    \centering
    \includegraphics[width=0.8\linewidth]{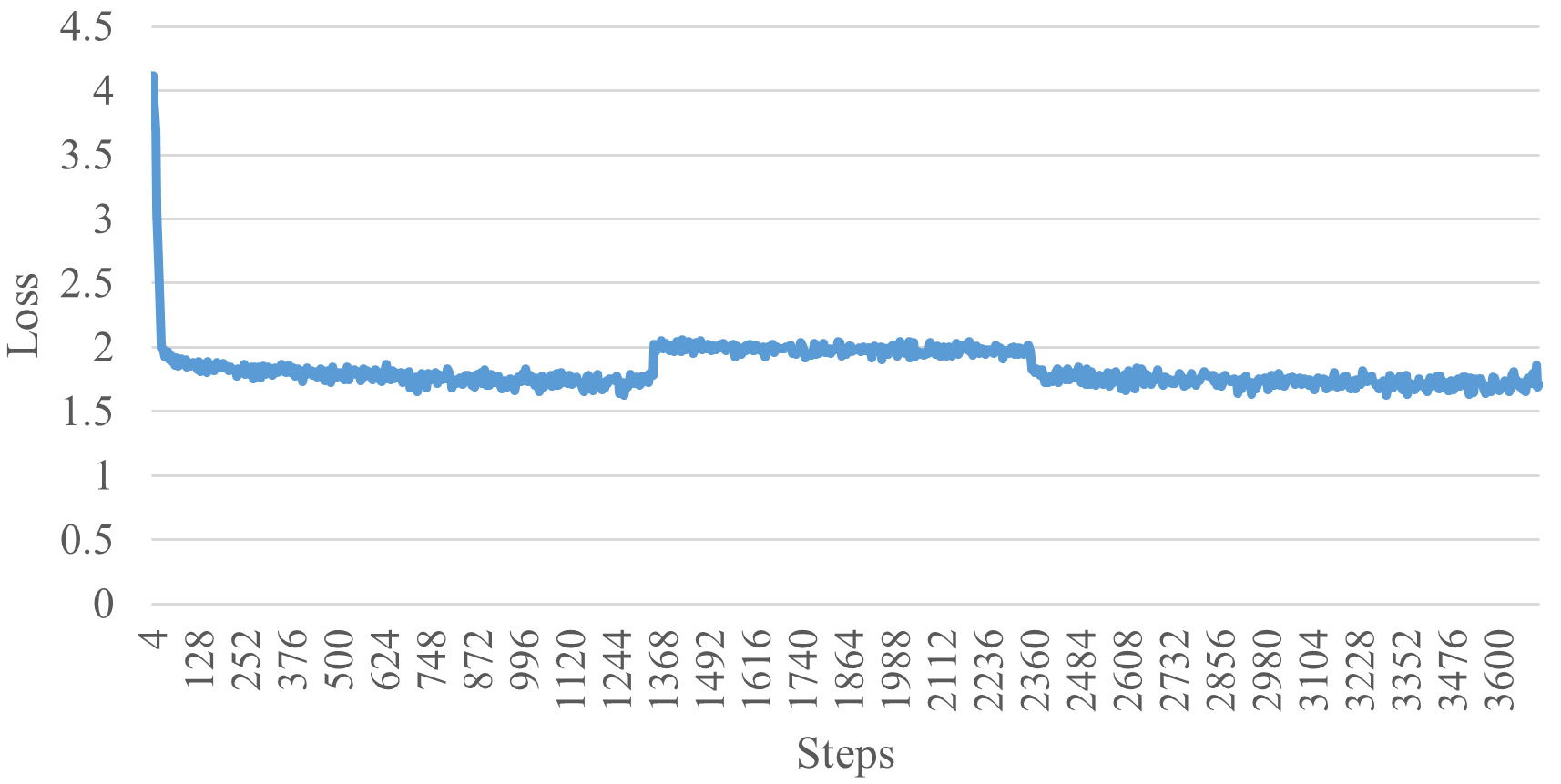}
    \caption{Training loss curve during the fine-tuning stage.}
    \label{fig3}
\end{figure}

This streamlined preparation, coupled with the LoRA-based fine-tuning, was further enhanced by Flash Attention 2 \cite{b30}.By minimizing memory overhead, Flash Attention 2 allowed us to handle longer context lengths and larger batch sizes efficiently, enabling effective fine-tuning for the next-token prediction objective while balancing computational efficiency with model performance.
Figure \ref{fig3} presents the loss curve during the fine-tuning stage. The curve demonstrates a smooth and consistent decrease in training loss, indicating stable convergence without signs of divergence or severe overfitting.

	\subsection{Instruction tuning}
Following the fine-tuning stage, we performed instruction-tuning on the fine-tuned model using our crawled medical free-form Farsi question-answering (MF3QA) dataset.As you can see in table \ref{tab:ft_it_training_detail} this stage utilized the default template of the aya-expanse model and retained the same techniques and almost the same hyperparameters as the fine-tuning stage, with a few adjustments. These adjustments were made to prevent the model from overfitting on the MF3QA dataset and to ensure it primarily learned the speaking style of a doctor. Specifically, we employed the LoRA method with a rank of 2, an alpha value of 2, and a dropout rate of 0.4, while increasing the weight decay to 0.5 (instead of 0.1). The instruction-tuning process was conducted for a single epoch, enabling the model to better understand and generate responses tailored to Farsi question-answering tasks. This targeted optimization further refined the model’s capabilities, enhancing its effectiveness on our specific dataset.
Figure \ref{fig5} illustrates the training loss during instruction-tuning. Due to the smaller dataset size and single-epoch training, the loss converges more quickly, reflecting the model’s adaptation to the task-specific speaking style without overfitting.

	\begin{table}[ht]
		\centering
		\caption{Training hyperparameters} 
		\begin{tabular}{|l|c|c|}  
			\hline
			& Fine-tuning & Instruction-tuning \\ \hline
			Number of epochs & 1& 1\\ \hline
			Batch size & 2 &  2 \\ \hline
			Number of gradient &  &  \\ 
                        accumulation steps &16  &16 \\ \hline
                        Optimizer & AdamW & AdamW\\ \hline
                        Learning rate &5E-4 & 5E-4 \\ \hline
			Maximum gradient norm& 0.3 & 0.3 \\ \hline
			Warmup ratio&0.03 &0.03	\\ \hline
                        Weight decay rate & 0.1  &0.5	\\ \hline
                        Maximum context length & 1024  &1024	\\ \hline
                        Padding strategy & left side padding   & left side padding	\\ \hline
                        Lora rank &8   &2	\\ \hline
                        Lora alpha &16   &2	\\ \hline
                        Lora dropout rate &0.05   &0.4	\\ \hline
                        Target modules &all linear layers   &all linear layers	\\ \hline
		\end{tabular}
		\label{tab:ft_it_training_detail}
	\end{table}

\begin{figure}[h]
    \centering
    \includegraphics[width=0.8\linewidth]{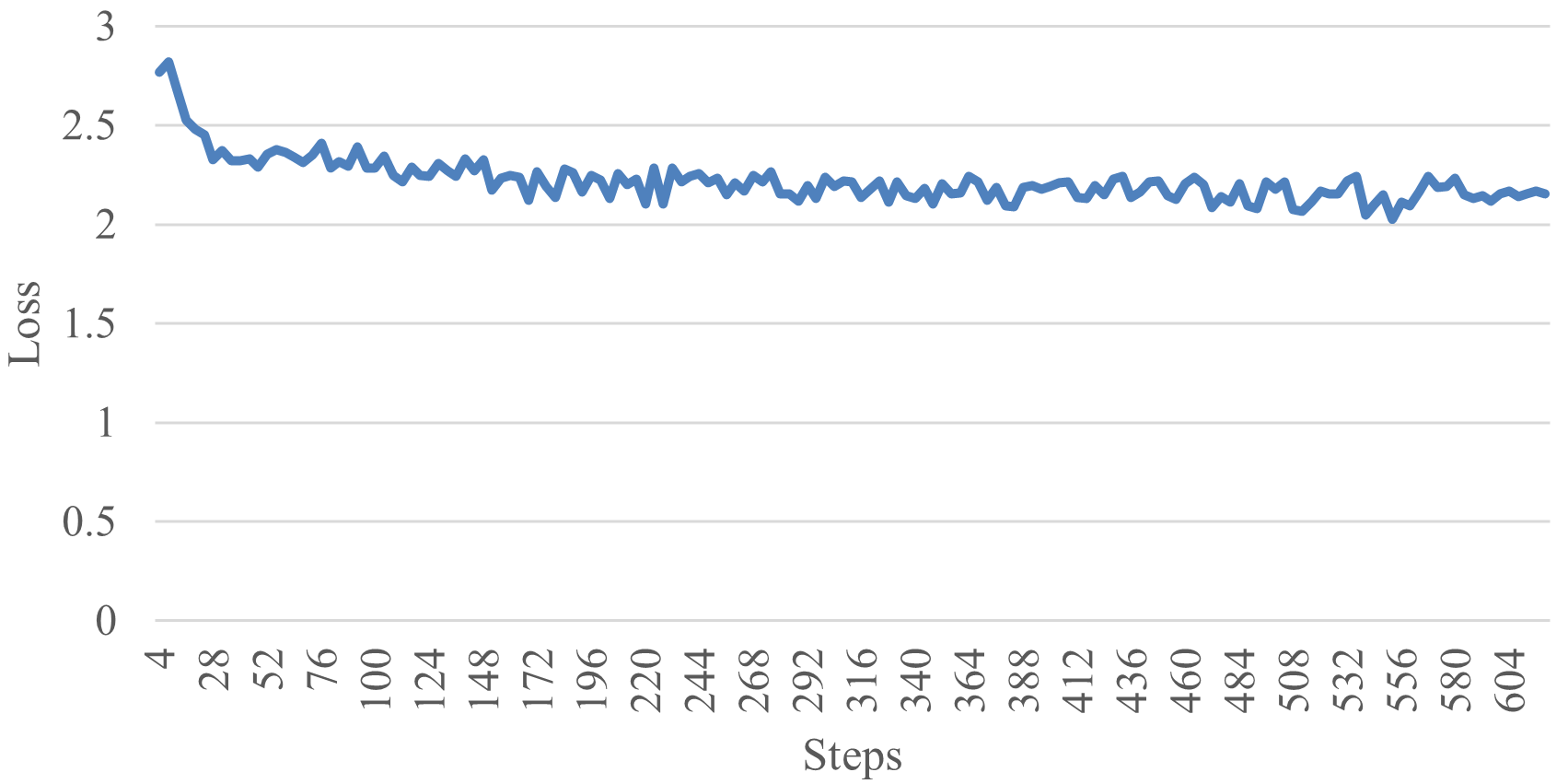}
    \caption{Training loss curve during the instruction-tuning stage.}
    \label{fig4}
\end{figure}

	\subsection{Carbon Footprint}
The carbon footprint of our model optimization—including both fine-tuning and instruction-tuning—was estimated based on hardware specifications and operational duration. The process ran for a combined total of 19 hours on a NVIDIA A100 PCIe 40GB GPU hosted in Google Cloud Platform’s asia-east1 region, utilizing a Google Colab Pro+ account for the hardware resources. Assuming a typical power consumption of 250 watts per GPU, the total energy used was 4.75 KWh (250 watts × 19 hours). Using the carbon intensity factor of the asia-east1 grid (0.56 kilograms of CO2 equivalent per kWh), this translates to 2.66 kilograms of CO2 equivalent emitted during the tuning process \cite{b31}.
	\section{Results}
	In the absence of a publicly available Persian medical language model, we opted to evaluate our model against general-purpose language models to establish a baseline for performance. This comparison allows us to assess the efficacy of our specialized model in handling medical-related queries in Persian. Importantly, all models used for comparison were selected based on their suitability for small, runnable environments on home devices, addressing privacy concerns prevalent in the medical domain. Additionally, we compared our model with a pipeline alternative in our evaluation, which consists of a series of processes: first, a translator converts the user’s query from Persian to English; next, this English query is input into an English medical language model; and finally, the response generated by the English model is translated back from English to Persian. By contrasting our model with both general-purpose language models and this pipeline alternative, we aim to demonstrate the advantages and specific capabilities of our small Persian medical language model in addressing the unique challenges of medical language processing within the Persian language context. 
	\footnote{You can see the detailed result at https://github.com/Mehrdadghassabi/Gaokerena-V}

Notice that all experiments were conducted on an NVIDIA L4 GPU, which is available through the Google Colab Pro and Pro+ plans. Leveraging this hardware ensured that the experiments were carried out in an accessible yet capable computational environment, reflecting practical limitations similar to those faced by many researchers.
	
	Moreover, we also address the significant challenge posed by the lack of available Persian benchmarks for medical language processing. To overcome this challenge, we translated the medical portion of the Massive Multitask Language Understanding (MMLU) dataset 
	\cite{b32} 
	into Persian and supplemented it with data from the Iranian Basic Medical Sciences Entrance Exam (IBMSEE).

\subsection{Comparison with general purpose language models}
As shown in Table \ref{tab:model_results_vs_general_purpose_languages}, our model achieved remarkable success by surpassing the passing score of the Iranian Basic Medical Sciences Entrance Exam, which is set at 36\%. This achievement makes it the first Persian language model with fewer than 8 billion parameters to pass this exam. Furthermore, our model demonstrated significant improvements on the translated MMLU dataset, achieving higher average scores and excelling across most sub-categories. This highlights its effectiveness in understanding and generating medical knowledge in the Persian language.

Notably, in the anatomy category, the model achieved a 7.4\% improvement compared to the baseline model. This improvement is likely due to the availability of more anatomy-related data in authoritative online magazines compared to other categories. However, it also underscores the lack of publicly available medical data in the Persian language, which poses challenges for further enhancing the model’s performance across other medical subcategories.

Another key observation, as shown in Table \ref{tab:model_results_vs_general_purpose_languages}, is that PersianMind exhibits low inference time, primarily because it generates very short responses compared to other models, producing the end-of-sentence token much sooner.

\begin{figure}[htbp]
		\centerline{\includegraphics[width=0.48\textwidth]{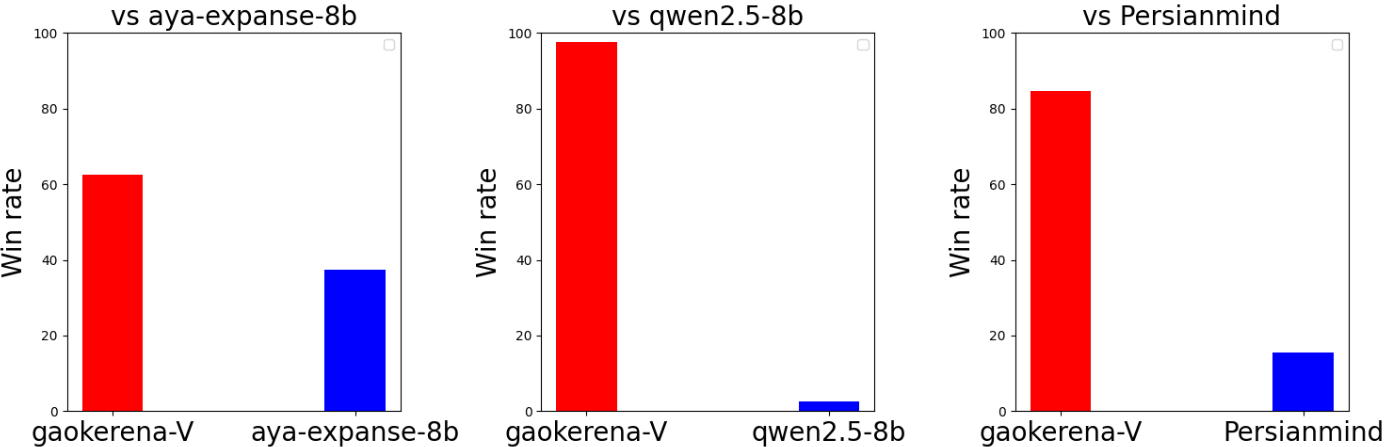}}
		\caption{Our model win rate against general purpose language models}
		\label{fig5}
	\end{figure}

In addition to our multiple-choice question-answering evaluation, we utilized GPT-4o \cite{b33} as an evaluator for free-form question answering. We provided the test set from the MF3QA dataset, which includes 2,000 records, to both our model and competing language models. As shown in Figure \ref{fig5}, GPT-4o predominantly preferred the responses generated by our model over those from the other three language models tested. This demonstrates that our model delivers high-quality responses, as judged by an advanced language model.

Moreover, GPT-4o was prompted to prefer responses that provide more accurate medical information while avoiding incorrect medical information. This further validates the reliability and precision of our model in generating medically accurate answers.
\footnote{you can see the prompt at https://github.com/Mehrdadghassabi/Gaokerena-V/blob/main/evaluation/free\_form\_qa/gaokerena\_vs\_aya/Untitled20.ipynb}

	\begin{table}[ht]
		\centering
		\caption{
			our model performance 
			in comparison with general purpose language models
		}
		\begin{tabular}{|l|c|c|c|c|}  
			\hline
			\textbf{} & \textbf{gao} & \textbf{aya-} &  &  \\ 
			& \textbf{kerena-V} & \textbf{expanse-8b} & \textbf{Qwen2.5} & \textbf{PersianMind} \\
			& (ours) & (baseline) &  &  \\ \hline
			MMLU- & \textbf{48.14} & 40.74 & 41.48 & 25.18 \\ 
			anatomy(fa) &  &  &  &  \\ \hline
			MMLU- &  &  &  &  \\
			medical & \textbf{53.0} & 49.0 & 52.0 & 34.0 \\ 
			genetics(fa) &  &  &  &  \\ \hline
			MMLU- &  &  &  &  \\
			college & 43.93 & \textbf{44.51} & 43.35 & 20.23 \\
			medicine(fa) &  &  &  &  \\ \hline
			MMLU- &  &  &  &  \\
			clinical& \textbf{55.47} & 52.07 & 47.92 & 25.28 \\
			knowledge(fa)&  &  &  &  \\ \hline
			MMLU- &  &  &  &  \\
			professional& \textbf{47.05} & 45.58 & 43.01 & 23.89 \\ 
			medicine(fa)&  &  &  &  \\ \hline
			MMLU- &  &  &  &  \\
			college& \textbf{47.22} & 45.14 & 42.36 & 32.63 \\
			biology(fa)&  &  &  &  \\ \hline
			MMLU(avg) & \textbf{49.31} & 46.64 & 45.17 & 25.89 \\ \hline
			IBMSEE Sept &  &  &  &  \\ 
			2023 & \textbf{38.69} & 34.52 & 33.33 & 19.64 \\  \hline
			Number of&  &  &  &  \\
			parameters & 8b & 8b & 7.6b & 6.8b \\ \hline
			inference time & $\approx 10s$ & $\approx 10s$ & $\approx 15s$ & $\approx 2s$ \\  \hline

		\end{tabular}
		\label{tab:model_results_vs_general_purpose_languages}
	\end{table}
	
	\subsection{Comparison with pipeline alternatives}
	As previously mentioned, one alternative to creating a Persian medical language model is a pipeline system. However, a major problem with pipeline systems is their speed; they exhibit high inference times because the output from one model must be fed into a second model, and then the output of the second model is processed again by the first model. This iterative process significantly hampers efficiency. To address the low speed of pipeline models, we have loaded all parameters—both those pertaining to the translators and the medical language model—simultaneously.

Our experiments with models such as Medmobile \cite{b10} paired with gemma-2b-it \cite{b15} as translators, and Medmobile paired with Parsinlu \cite{b33} \cite{b35} models, showed disappointing results, as evidenced by the poor performance displayed in Table \ref{tab:model_results_vs_pipeline_alternative}. Since the translators used in these experiments were not specifically designed for medical translation and were quite small, making them suitable for running on home devices, this approach yielded poor results, nearing random choice in the MMLU and IBMSEE evaluations. Thus, we concluded that this alternative is not viable, and we did not examine any other translators.

Another significant issue with the pipeline alternative is its poor performance in accurately detecting and translating medical terms. This limitation poses a serious challenge, as precise terminology is crucial for effective communication in healthcare settings, where misunderstandings can have serious consequences for patient care and treatment outcomes. The underlying cause of this deficiency is likely due to the fact that the translators employed in these pipeline systems have not been specifically developed for medical translation. Unlike general-purpose translation models, medical translation requires a nuanced understanding of specialized vocabulary, context, and the intricacies of medical language.

\begin{figure}[htbp]
		\centerline{\includegraphics[width=0.48\textwidth]{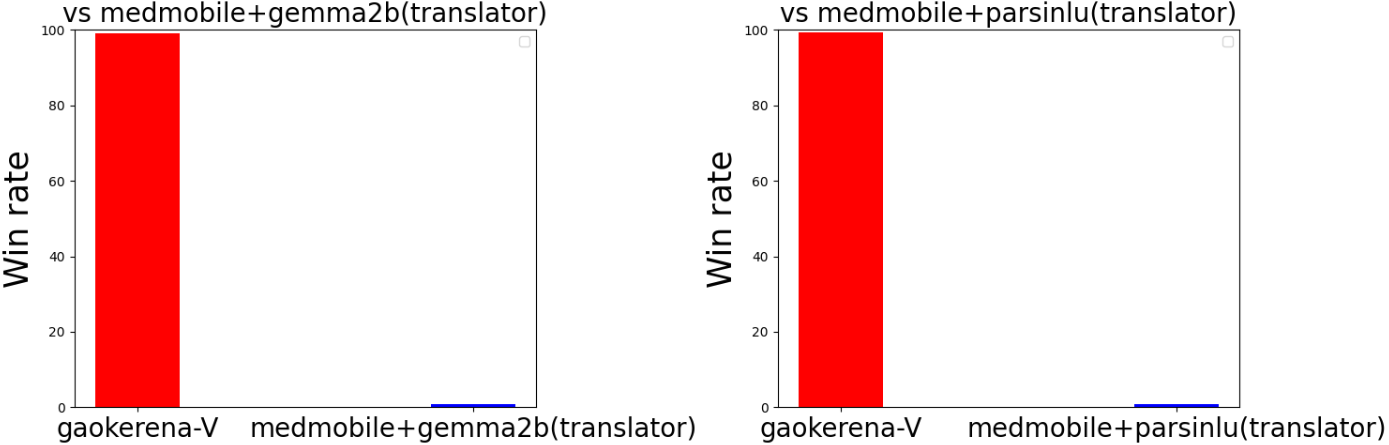}}
		\caption{Our model win rate against pipeline alternatives}
		\label{fig6}
	\end{figure}

Currently, there are no models tailored for medical translation in the Persian language, which means that existing systems are ill-equipped to handle the complexities of medical terminology. As shown in Figure \ref{fig6}, these limitations have resulted in all pipeline alternatives achieving lower win rates against our model, Gaokerena-V. This performance gap highlights the inadequacies of current pipeline approaches in meeting the demands of medical translation, emphasizing the necessity for the development of dedicated medical translator models that can effectively address these specific challenges.
	
	\begin{table}[ht]
		\centering
		\caption{our model performance 
			in comparison with pipeline alternatives}
		\begin{tabular}{|l|c|c|c|}  
			\hline
			\textbf{} & \textbf{gao} 
			& \textbf{MedMobile} & \textbf{MedMobile} \\ 
			& \textbf{kerena-V} & + \textbf{gemma2} & \textbf{+ parsinlu} \\
			& (ours)  & \textbf{-2b-it} &  \\ \hline
			MMLU- &  &  &  \\ 
			anatomy(fa)  & \textbf{48.14} & 14.07 & 25.18  \\ \hline
			MMLU- &    &  &  \\
			medical-genetics(fa) & \textbf{53.0} & 20.0 & 35.0 \\ \hline
			MMLU- &  &    &  \\
			college-medicine(fa) & \textbf{43.93} & 19.08 & 27.17 \\ \hline
			MMLU- &    &  &  \\
			clinical-knowledge(fa)& \textbf{55.47} & 27.54 & 31.70 \\ \hline
			MMLU- &  &  &  \\
			professional-& \textbf{47.05} & 17.27 & 33.82 \\
                   medicine(fa)& &  &  \\ \hline
			MMLU- &  &  &  \\
			college-biology(fa)& \textbf{47.22} & 18.75 & 31.25 \\ \hline
			MMLU(avg) & \textbf{49.31} & 20.11 & 30.99 \\ \hline
			IBMSEE Sept2023 & \textbf{38.69}  & 24.40 & 32.73  \\ \hline
			Number of&  &  &  \\
			parameters & 8b & 3.8b+2b & 3.8b+1.2b+1.2b \\ \hline
			inference time & $\approx 10s$ & $\approx 20s$ & $\approx 30s$ \\  \hline
		\end{tabular}
		\label{tab:model_results_vs_pipeline_alternative}
	\end{table}
	
	\section*{Future research}
	Although Gaokerena-V demonstrates enhanced Persian medical knowledge compared to its baseline and other alternatives, its current performance is not yet sufficient for reliable use in clinical scenarios. As mentioned in the baseline model section, we plan to merge our updated    parameters into aya-vision rather than aya-expanse, enabling the model to incorporate multimodal inputs such as medical images (e.g., MRIs and CT scans). However, the model’s capabilities in handling multimodal queries remain under-explored. Our research will focus on improving its ability to interpret and align textual and visual information, with the goal of expanding its diagnostic and analytical utility in medicine. Additionally, future work will involve collaboration with real doctors to enhance the model through reinforcement learning from human feedback (RLHF)
\cite{b36} and rigorous clinical validation, ensuring its practical applicability and safety in real-world healthcare settings.

\end{document}